\documentclass[11pt,letterpaper,twocolumn]{article}

\usepackage[letterpaper,top=0.75in,bottom=0.75in,left=0.75in,right=0.75in,columnsep=0.25in]{geometry}
\usepackage[T1]{fontenc}
\usepackage{times}
\usepackage{amsmath}
\usepackage{newtxmath}
\usepackage{bm}
\usepackage{graphicx,booktabs,array}
\usepackage{microtype}
\usepackage{cite}
\usepackage{xcolor}
\definecolor{linkblue}{HTML}{416F9F}
\usepackage[breaklinks,colorlinks,allcolors=linkblue]{hyperref}
\hypersetup{pdftitle={Sparse Text Diffusion under Dual Budgets: Context Connectivity and Expert Activation},pdfauthor={Alexandros Christoforos; Chadbourne Davis; Paul Raj}}
\newcommand{\method}{DualBudget}
\newcommand{\Nset}{\mathcal{N}}
\newcommand{\Gset}{\mathcal{G}}
\newcommand{\Sset}{\mathcal{S}}

\newcommand{\SoftmaxOp}{\operatorname{softmax}}
\newcommand{\MoEOp}{\operatorname{MoE}}
\newcommand{\norm}[1]{\lVert #1\rVert}
\makeatletter
\renewcommand\section{\@startsection{section}{1}{0pt}
  {-2ex plus -.4ex minus -.2ex}{.8ex plus .2ex}{\normalfont\large\bfseries}}
\renewcommand\subsection{\@startsection{subsection}{2}{0pt}
  {-1.4ex plus -.3ex minus -.1ex}{.5ex plus .1ex}{\normalfont\normalsize\bfseries}}
\makeatother

\title{Sparse Text Diffusion under Dual Budgets:\\Context Connectivity and Expert Activation}

\author{Alexandros Christoforos\textsuperscript{1}\qquad
Chadbourne Davis\textsuperscript{1}\qquad
Paul Raj\textsuperscript{2}\\[0.4em]
\normalsize\textsuperscript{1}Suffolk University\\
\normalsize\textsuperscript{2}Texas A\&M University-Commerce}

\date{}
\begin{document}
\maketitle

% Source: sections/00_abstract
\begin{abstract}
Long-context diffusion repeatedly pays for both token interaction and feed-forward computation. We formulate these costs as two separate budgets: the edges available to an attention graph and the experts activated for each token. This view connects local, dilated, and global attention with conditional expert computation without assuming that sparsity alone preserves document-level information. We specify normalized masked attention, distinguish denoising depth from sampling iterations, and derive graph and compute bounds. A reported sparse-expert configuration, denoted DualBudget, is analyzed across scientific summarization, multi-hop question answering, dialogue, and sentence-pair prediction. All supplied table measurements are retained. The length study shows gains over the listed baselines through 16K, together with measurable degradation as length increases. Ablations reveal mixed changes across the three reported scores, while the cost record indicates a 20\% reduction relative to DiffuSeq under an unspecified timing convention. The analysis separates these observations from unmeasured claims about training speed, absorbing states, global coherence, and deployment memory.
\end{abstract}

% Source: sections/01_introduction
\section{Introduction}
Denoising diffusion and score-based modeling provide iterative generative principles~\cite{ho2020ddpm,song2021sde}. DiffuSeq formulates sequence-to-sequence generation~\cite{gong2022diffuseq}, with Transformer attention providing interactions within each denoiser~\cite{vaswani2017attention}. Longer sequences increase the cost of every update: dense query--key interactions grow quadratically, while expanded feed-forward capacity remains expensive even under restricted attention.

These costs motivate joint accounting of \emph{context connectivity} and \emph{expert activation}. Longformer combines local and task-motivated global attention~\cite{beltagy2020longformer}. Sparse expert layers select which parameterized functions process each token~\cite{shazeer2017,fedus2022switch}. The former controls information access; the latter controls conditional computation. Reducing one need not reduce the other: a short neighborhood can feed an expensive dense sublayer, while sparse experts can consume representations built through dense attention.

Related mechanisms already coexist. DrDiff combines expert scheduling, hierarchical sparse attention, and accelerated sampling for long-text diffusion~\cite{req_drdiff}; DiffuSeq-v2 introduces soft absorbing states and continuous-space solvers~\cite{gong2023diffuseq}. We formulate common budgets to analyze the reported sparse-expert comparison, rather than claim these component families originate here. The formulation exposes separate requirements: attention must preserve useful paths, routing must specify activated experts and dispatch costs, and sampling must identify its state and prediction convention.

At the task level, agents also allocate scarce reasoning capacity. Markets of Minds evaluates expert assistance through marginal benefit and opportunity cost under shared budgets~\cite{req_markets}. This motivates an agent-facing interpretation of the two budgets: a planner can assign priorities to document spans, while the denoiser determines which token connections and expert activations realize them. Span-level coordination defines an extension interface; the experiments below concern the token-level denoiser.

We develop this perspective in three stages. First, we formalize a sparse attention graph and a selected-expert layer with explicit normalization. Second, we derive interaction and active-compute bounds and explain their connectivity and memory limits. Third, we analyze the supplied measurements, retaining aggregate scores, length sweeps, ablations, and cost values as separate records. The resulting account identifies which efficiency and quality conclusions follow from those records and which require additional instrumentation.

% Source: figures/architecture
\begin{figure*}[t]
\centering
\includegraphics[width=\textwidth]{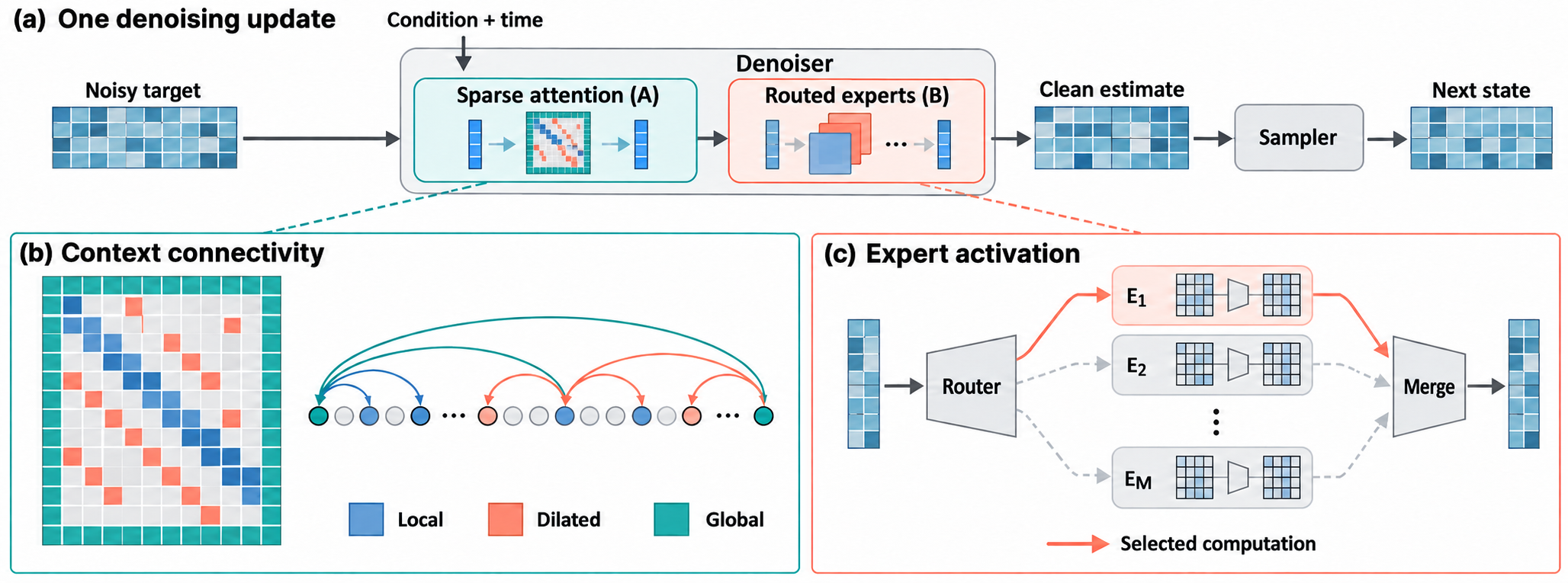}
\caption{One denoising evaluation and its two computational budgets. (a) The denoiser predicts a clean state for the sampling update. (b) Sparse token interactions. (c) Selected expert computation. Highlighting is schematic and specifies no measured activation count.}
\label{fig:architecture}
\end{figure*}

% Source: sections/02_related
\section{Related Work}
\subsection{Diffusion Representations and Sampling}
Diffusion-LM denoises continuous word representations, while SSD-LM generates text semi-autoregressively in simplex space~\cite{li2022diffusionlm,han2022ssdlm}. Structured discrete transitions~\cite{austin2021structured}, score-entropy estimation in SEDD~\cite{lou2024sedd}, and masked objectives in MDLM~\cite{sahoo2024mdlm} offer different treatments of discrete language. LLaDA extends masked diffusion to large-scale language modeling~\cite{nie2025llada}; soft-masked corruption provides another training design~\cite{chen-etal-2023-cheaper}. These choices concern representations and objectives, without fixing attention connectivity or expert dispatch.

Counterfactual adaptation in CF-VLM and semantic caption refinement~\cite{req_cfvlm,req_tdsr} operate at other levels. Autoregressive diffusion explores prediction decomposition~\cite{hoogeboom2021autoregressive}; DPM-Solver++ specifies accelerated numerical integration~\cite{lu2022dpm}. Failure-driven workflow refinement concerns procedural organization~\cite{req_workflow}. Their distinct roles motivate separating corruption, computation, and scheduling costs.

\subsection{Expert Routing and Execution}
GShard and GLaM scale conditional capacity~\cite{lepikhin2020gshard,du2022glam}, while ST-MoE examines stability and transfer~\cite{zoph2022stmoe}. KABB coordinates agent expertise~\cite{req_kabb}. Expert-choice routing changes capacity assignment~\cite{zhou2022mixture}; MegaBlocks addresses dynamic dispatch through block-sparse execution~\cite{gale2022megablocks}. GAM-Agent studies visual-reasoning collaboration~\cite{req_gam}. DeepSpeed-MoE targets system execution~\cite{rajbhandari2022deepspeed}, whereas MAT-Agent concerns training coordination~\cite{req_mat}.

Parameter-efficient expert architectures~\cite{gao2022parameter}, Mixtral~\cite{jiang2024mixtral}, and DeepSeekMoE~\cite{dai2024deepseekmoe} illustrate adaptation and specialization choices. OSC aligns collaborator knowledge and Agora exchanges uncertainty~\cite{req_osc,req_agora}. Token dispatch, expert residency, and agent assistance consequently require different resource accounting.

\subsection{Context Mechanisms and Evaluation Scope}
Sparse Transformers, BigBird, and ETC construct restricted interactions~\cite{child2019sparse,zaheer2020big,ainslie-etal-2020-etc}. Reformer uses hashing~\cite{kitaev2020reformer}; Linformer and Performer use low-rank and feature-based approximations~\cite{wang2020linformer,choromanski2020performer}. FlashAttention improves IO efficiency for exact attention~\cite{dao2022flash}. These mechanisms should be distinguished from changing the explicit mask alone.

HTC-VLM and image-based video encoders alter visual-token processing~\cite{req_htc,req_video}. DepthSSC adapts geometric representations~\cite{req_depthssc}; Tri-MARF and MM-CoT address annotation and visual reasoning~\cite{req_tri,req_mmcot}. The NTIRE restoration challenge~\cite{req_ntire}, ActPhysCause video report~\cite{req_actphys}, and deepfake scaling study~\cite{req_deepfake} provide cross-domain evaluation contexts. Their metrics and budgets do not substitute for text-diffusion measurements.

\subsection{Task-Level Agent Coordination}
ReAct interleaves reasoning and actions~\cite{yao2022react}; AutoGen supports multi-agent conversations~\cite{wu2023autogen}. CAMEL studies role-based cooperation~\cite{li2023camel}, while multiagent debate refines responses through repeated exchanges~\cite{du2023debate}. These supply context for a task-level coordinator whose requests would consume denoiser resources. That extension is distinct from the evaluated token-level configuration.

\subsection{Closest Diffusion Comparison}
DrDiff overlaps directly with sparse expert denoising, but is absent from the supplied score tables. Its routing, masks, and solver are not matched here, so no numerical superiority is claimed. The present analysis makes graph accounting, prediction conventions, and the available comparison's empirical scope explicit.

% Source: sections/03_formulation
\section{Dual-Budget Denoising}
\subsection{States, Time, and Network Depth}
Let $y=(y_1,\ldots,y_n)$ be target tokens, $c$ the conditioning input, and $z_0=\operatorname{Emb}(y)\in\mathbb R^{n\times h}$ their clean embeddings. Diffusion time $t$ indexes corruption and reconstruction; layer index $\ell$ indexes computation inside a denoiser. These indices are independent. A model with $L$ layers evaluated for $S$ sampling updates incurs $S$ denoiser evaluations, not a single $L$-step diffusion trajectory.

For a Gaussian marginal, write
\begin{equation}
z_t=\sqrt{\bar\alpha_t}\,z_0+
\sqrt{1-\bar\alpha_t}\,\epsilon,
\quad \epsilon\sim\mathcal N(0,I).
\label{eq:forward}
\end{equation}
The denoiser receives the noisy target, conditioning information, and time. We use clean-state prediction notation
\begin{equation}
\hat z_0=f_\theta(z_t,t,c).
\label{eq:predict}
\end{equation}
This notation fixes the object being predicted; it does not identify an unreported checkpoint parameterization. Predicting noise or the previous state requires its corresponding conversion and training objective.

\subsection{The Context Budget as an Attention Graph}
At a layer, let $Q,K,V$ be projected token representations. For query position $i$, define an allowed key set $\Nset_i$. Local or dilated neighborhoods are generated from an offset set $\mathcal W$ of at most $w$ offsets and a stride $\delta$:
\begin{equation}
\mathcal W_i=\{i+\delta r:r\in\mathcal W\}\cap\{1,\ldots,n\}.
\label{eq:window}
\end{equation}
Let $\Gset$ contain $g$ designated global positions. The combined neighborhood is
\begin{equation}
\Nset_i=
\begin{cases}
\{1,\ldots,n\},&i\in\Gset,\\
\mathcal W_i\cup\Gset,&i\notin\Gset.
\end{cases}
\label{eq:neighborhood}
\end{equation}
The graph edge $(i,j)$ denotes query $i$ attending to key $j$. Global queries read all positions, while ordinary queries can read global keys. A causal task would additionally intersect the allowed set with its causal mask; the full bidirectional pattern is not automatically valid for every prediction interface.

Attention is normalized over the entire allowed key set:
\begin{equation}
\begin{aligned}
a_{ij}&=\frac{\exp(q_i^\top k_j/\sqrt{d_k})}
 {\sum_{u\in\Nset_i}\exp(q_i^\top k_u/\sqrt{d_k})},\\
o_i&=\sum_{j\in\Nset_i}a_{ij}v_j.
\end{aligned}
\label{eq:attention}
\end{equation}
This definition avoids applying a scalar softmax independently to a single query--key pair. It also separates the mask from the learned attention values: an allowed edge can receive little weight, and a high-value excluded edge remains unavailable.

% Source: figures/connectivity
\begin{figure}[t]
\centering
\includegraphics[width=\columnwidth]{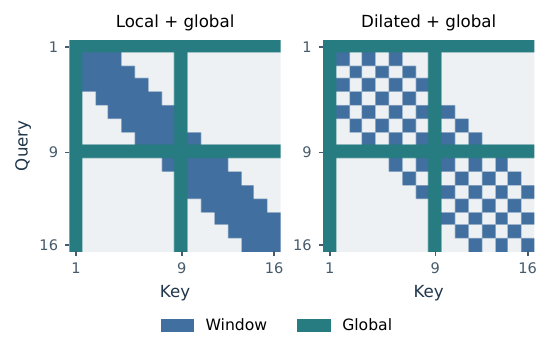}
\caption{Constructed attention masks illustrating local and dilated access with global positions. Rows are queries and columns are keys. These are protocol illustrations, not learned masks or benchmark measurements.}
\label{fig:connectivity}
\end{figure}

\subsection{The Expert Budget as Conditional Computation}
Let $u_i$ be the representation entering an expert layer and $E_1,\ldots,E_M$ its expert functions. A router produces probabilities
\begin{equation}
p_i=\SoftmaxOp(W_r u_i),\qquad \Sset_i\subseteq\{1,\ldots,M\}.
\label{eq:router}
\end{equation}
The selected set $\Sset_i$ defines which experts execute. Its size is bounded by an activation budget $k$. A general selected-expert output is
\begin{equation}
\MoEOp(u_i)=\sum_{e\in\Sset_i}\pi_{ie}E_e(u_i),
\quad |\Sset_i|\leq k.
\label{eq:moe}
\end{equation}
The weights $\pi_{ie}$ may use selected router probabilities with or without renormalization. The choice affects the numerical function and must be recorded. A dense softmax over all experts does not establish sparse execution unless a selection or dispatch rule is also specified.

Token selection is distinct from expert residency. An implementation may execute only $k$ experts for one token while storing all $M$ experts and their optimizer state. Capacity limits, dropped or rerouted tokens, load balancing, and dispatch communication can further change both output quality and wall-clock cost. We keep these implementation quantities explicit rather than infer them from the word ``sparse.''

\subsection{Training and the Sampling Interface}
Under clean-state prediction, a reconstruction objective can be written
\begin{equation}
\mathcal L_{\rm rec}=\mathbb E_{t,\epsilon,y,c}
 \norm{z_0-f_\theta(z_t,t,c)}_F^2+R(z_0).
\label{eq:loss}
\end{equation}
Here $R$ denotes the latent regularization described in the source formulation. Its functional form and coefficient are not specified by the score tables. Token position must not be conflated with diffusion time in the reconstruction target.

Inference requires a sampler compatible with the trained prediction convention. We write its interface as
\begin{equation}
z_{t'}=\Psi\bigl(z_t,f_\theta(z_t,t,c);t,t'\bigr),\quad t'<t.
\label{eq:sampler}
\end{equation}
For a continuous formulation with velocity $v_\theta$, the associated integral is $z_t=z_s+\int_s^t v_\theta(z_\tau,\tau,c)\,d\tau$. A denoiser output cannot simply be exponentiated and substituted for this velocity. DDIM changes the sampling path while retaining the training objective~\cite{song2020ddim}; Euler and DPM-Solver++ use different numerical updates. Because the record mentions both without resolving the implementation, the reported comparison is not assigned to one particular solver here.

Absorbing or soft-masked corruption is a separate design choice. DiffuSeq-v2 already introduces a soft absorbing state~\cite{gong2023diffuseq}. The supplied comparison does not specify its transition schedule or isolate its contribution. We therefore retain it as an implementation question rather than attribute a measured stabilization benefit to an undefined mechanism.

% Source: sections/04_analysis
\section{Cost and Connectivity Analysis}
\subsection{Counting Token Interactions}
The context budget is the number of allowed attention edges, $B_{\rm ctx}=\sum_i|\Nset_i|$. Under Eq.~\eqref{eq:neighborhood},
\begin{equation}
B_{\rm ctx}\leq(n-g)(w+g)+gn.
\label{eq:edges}
\end{equation}
Each of the $n-g$ ordinary queries reads at most $w+g$ keys, while each global query reads $n$. Boundary clipping and overlaps can reduce the actual count. Thus attention-score and value-aggregation work scales as $O(n(w+g)h)$ for fixed-width projections, rather than $O(n^2h)$ for dense pairwise interactions. Query, key, and value projections still require their own matrix operations.

Linear scaling in $n$ requires $w$ and $g$ to remain bounded. If all tokens become global, the dense pattern returns. If window size grows with document length, the resulting complexity must include that growth. Dilation changes the spacing of selected positions, rather than their count, so it can expand geometric span without a proportional increase in allowed keys. This is a count statement, not a guarantee that all useful information remains reachable.

\subsection{Reachability and Global Bottlenecks}
For symmetric local windows with radius $r$ and fixed stride $\delta$, one layer moves information at most $\delta r$ positions along an available local edge. Across $L$ layers, a path's displacement is bounded by $L\delta r$. A large displacement bound does not imply that every intermediate position is reachable. For example, repeating stride two without other connections can preserve parity-separated components.

Global nodes change this argument. With both global-query and global-key edges, information from any position can reach a global position in one layer and another ordinary position in the next. The graph therefore admits short paths. The transmitted representation still has finite width and learned attention weights, so short paths do not establish lossless document storage or guaranteed factual coherence. Position-sensitive context use remains an empirical concern~\cite{liu2023lostmiddle}; a small graph diameter alone does not resolve it.

The toy masks in Figure~\ref{fig:connectivity} illustrate these structural differences. They are constructed examples, not learned masks or benchmark observations. Their role is to make explicit the connectivity assumptions behind an asymptotic claim. A measured implementation should separately record mask degree, selected global positions, and the distribution of attention mass.

\subsection{Combining Context and Expert Budgets}
Define the expert-activation budget as $B_{\rm exp}=\sum_i|\Sset_i|\leq nk$. If one expert has hidden width $h_f$, its feed-forward matrix work is proportional to $hh_f$. Across $L$ layers and $S$ denoiser evaluations, a useful decomposition is
\begin{equation}
\begin{aligned}
C_{\rm total}=SL\bigl[&O(nh^2)+O(B_{\rm ctx}h)\\
 &+O(B_{\rm exp}hh_f)+C_{\rm route}\bigr].
\end{aligned}
\label{eq:cost}
\end{equation}
The terms represent projection work, sparse interactions, selected experts, and routing/dispatch overhead. A conventional dense router can itself require work proportional to $nhM$. Communication and capacity padding depend on the execution system. Omitting these terms can make an activation-count reduction look like a wall-clock guarantee that it does not provide.

This decomposition also separates training corruption steps from inference evaluations. A training horizon of 2048 possible times does not imply that every inference sample uses 2048 denoiser calls. Conversely, reducing calls changes cost only through the implemented schedule and solver. Claims about fewer sampling steps should report the actual number of function evaluations and any quality change under a matched configuration.

\subsection{Memory and the Meaning of Efficiency}
Sparse masks reduce the score entries that need to be materialized by a compatible sparse kernel. They do not remove token representations, projections, expert parameters, gradients, or optimizer state. A dense implementation that masks a full score matrix may retain quadratic memory despite a sparse logical graph. Training and inference also have different residency requirements.

Practical efficiency therefore needs several measurements: peak allocated memory, stored parameter and optimizer state, attention workspace, dispatch buffers, denoiser evaluations, latency, and throughput. These quantities should be reported at matched batch size, precision, sequence length, and hardware. The available cost table supports a narrower within-record comparison; it does not provide a complete decomposition of Eq.~\eqref{eq:cost}.

% Source: figures/cost
\begin{figure}[t]
\centering
\includegraphics[width=\columnwidth]{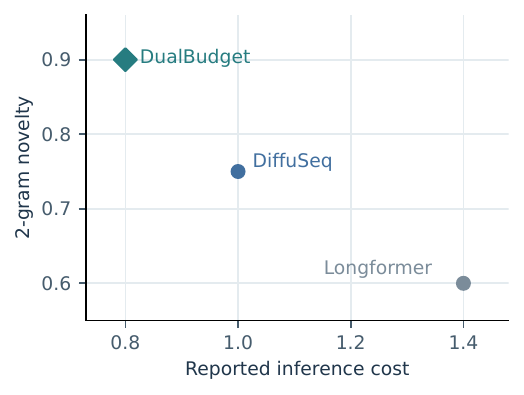}
\caption{Reported inference cost and 2-gram novelty. Coordinates reproduce the three source rows. The cost unit, normalization, and uncertainty are unspecified, so the plot supports an ordering within this record rather than a general latency claim.}
\label{fig:cost}
\end{figure}

% Source: sections/05_evaluation
\section{Empirical Evaluation}
\subsection{Tasks, Metrics, and Implementation Scope}
Scientific summarization uses the arXiv dataset associated with discourse-aware long-document summarization~\cite{cohan-etal-2018-discourse}. HotpotQA measures answer and supporting-fact prediction~\cite{yang-etal-2018-hotpotqa}. Commonsense-focused dialogue supplies response-generation examples~\cite{zhou-etal-2021-commonsense}. QQP concerns sentence-pair decisions; the cited bilateral matching model is background for sentence matching~\cite{wang2017bilateral}. Its accuracy is not a direct test of generated paraphrase quality.

ROUGE~\cite{lin2004rouge}, BLEU~\cite{papineni2002bleu}, and BERTScore~\cite{zhang2019bertscore} capture different similarities. The dialogue table retains fractional scores, while the summarization table reports ROUGE values on a larger numeric scale. We preserve those scales and avoid combining them into one overall quality score. None of these automatic measures alone establishes factual consistency, global discourse coherence, or causal reasoning.

The reported architecture has 12 layers and 12 attention heads, an A100 execution setting, a square-root noise schedule, and a 2048-step training horizon. The full ablation setting uses window size 512. The expert count, activation count, model width, global-token selection, dilation schedule, optimizer, data splits, and sampler implementation are unspecified. These omissions limit exact reproduction and attribution to the two budgets.

The comparison labels are retained at their documented scope. Longformer is a long-context architecture with a generative LED variant~\cite{beltagy2020longformer}; its decoder and checkpoint must be specified for a generation comparison. The table label DiffuSeq does not establish whether it uses the original model or its accelerated successor. GPT-4 appears in the source background~\cite{achiam2023gpt}, but no numerical GPT-4 row is available, so it is not described as an evaluated baseline.

% Source: tables/arxiv_performance
\begin{table}[t]
\centering
\small
\setlength{\tabcolsep}{4pt}
\begin{tabular}{lrrr}
\toprule
Model & R1 & R2 & RL\\
\midrule
Longformer & 41.44 & 17.52 & 38.70\\
DiffuSeq & 39.12 & 16.43 & 37.88\\
\textbf{DualBudget} & \textbf{44.41} & \textbf{18.73} & \textbf{39.89}\\
\bottomrule
\end{tabular}
\caption{Scientific summarization: aggregate ROUGE comparison. This record is separate from the length sweep.}
\label{tab:arxiv_performance}
\end{table}

% Source: tables/hotpotqa_performance
\begin{table}[t]
\centering
\small
\setlength{\tabcolsep}{4pt}
\begin{tabular}{lrr}
\toprule
Model & Answer EM/F1 & Support EM/F1\\
\midrule
Longformer & 71.21 / 82.42 & 65.11 / 89.50\\
DiffuSeq & 70.91 / 81.43 & 64.60 / 88.51\\
\textbf{DualBudget} & \textbf{72.88 / 85.42} & \textbf{66.69 / 90.40}\\
\bottomrule
\end{tabular}
\caption{Aggregate HotpotQA answer and supporting-fact scores.}
\label{tab:hotpotqa_performance}
\end{table}

% Source: tables/commonsense_performance
\begin{table}[t]
\centering
\small
\setlength{\tabcolsep}{4pt}
\begin{tabular}{lrrr}
\toprule
Model & BLEU & ROUGE-L & BERTScore\\
\midrule
Longformer & 0.030 & 0.139 & 0.602\\
DiffuSeq & 0.022 & 0.119 & 0.501\\
\textbf{DualBudget} & \textbf{0.049} & \textbf{0.233} & \textbf{0.628}\\
\bottomrule
\end{tabular}
\caption{Dialogue scores under the source evaluation. Values retain their original fractional scale.}
\label{tab:commonsense_performance}
\end{table}

% Source: tables/qqp_performance
\begin{table}[t]
\centering
\small
\setlength{\tabcolsep}{4pt}
\begin{tabular}{lr}
\toprule
Model & Accuracy\\
\midrule
Longformer & 92.3\\
DiffuSeq & 91.7\\
\textbf{DualBudget} & \textbf{95.3}\\
\bottomrule
\end{tabular}
\caption{QQP accuracy under the recorded protocol; this is not a direct measure of generated paraphrase quality.}
\label{tab:qqp_performance}
\end{table}

\subsection{Aggregate Task Results}
The sparse-expert row is denoted \method{} throughout the analysis. This descriptive label identifies the recorded sparse-expert configuration throughout the comparisons. On aggregate arXiv summarization, it records 44.41/18.73/39.89 for R1/R2/RL. Relative to Longformer, the differences are 2.97/1.21/1.19 points; relative to DiffuSeq they are 5.29/2.30/2.01. These are metric-specific improvements, rather than a single claim of a uniform 3--5 point gain.

For aggregate HotpotQA, answer EM/F1 reaches 72.88/85.42 and support EM/F1 reaches 66.69/90.40. Against Longformer, the four differences are 1.67, 3.00, 1.58, and 0.90 points. Stronger supporting-fact scores are consistent with improved evidence selection under the recorded task, but do not isolate an attention-mask mechanism. The models may differ in other training or architectural choices.

The dialogue row records 0.049 BLEU, 0.233 ROUGE-L, and 0.628 BERTScore. QQP accuracy is 95.3, compared with 92.3 and 91.7 for the two listed baselines. These results show breadth across the supplied records, but the tasks differ in target length and output structure. Sentence-pair accuracy and short dialogue response quality do not independently demonstrate long-document generation.

\subsection{Length Sensitivity through 16K}
The arXiv length sweep is separate from the aggregate table. At 8K, \method{} records 46.85/19.72/41.35; at 16K, it records 43.60/18.11/39.75. The corresponding decreases are 3.25, 1.61, and 1.60 points. The model retains higher scores than the listed baselines at each length, while its own scores still decline. We therefore describe relative resilience over the tested range rather than length-invariant generation quality.

% Source: tables/arxiv_performance_length
\begin{table*}[t]
\centering
\small
\setlength{\tabcolsep}{4pt}
\begin{tabular}{lrrrrrrrrr}
\toprule
Model & \multicolumn{3}{c}{8K} & \multicolumn{3}{c}{12K} & \multicolumn{3}{c}{16K}\\
 & R1 & R2 & RL & R1 & R2 & RL & R1 & R2 & RL\\
\midrule
Longformer & 43.22 & 18.34 & 39.98 & 42.10 & 17.78 & 39.20 & 40.05 & 16.94 & 38.10\\
DiffuSeq & 41.78 & 17.86 & 38.91 & 40.35 & 17.25 & 38.11 & 39.50 & 16.71 & 37.65\\
\textbf{DualBudget} & \textbf{46.85} & \textbf{19.72} & \textbf{41.35} & \textbf{45.40} & \textbf{18.94} & \textbf{40.50} & \textbf{43.60} & \textbf{18.11} & \textbf{39.75}\\
\bottomrule
\end{tabular}
\caption{Scientific summarization across reported context lengths. All three ROUGE metrics are retained.}
\label{tab:arxiv_performance_length}
\end{table*}

% Source: figures/length
\begin{figure*}[t]
\centering
\includegraphics[width=0.96\textwidth]{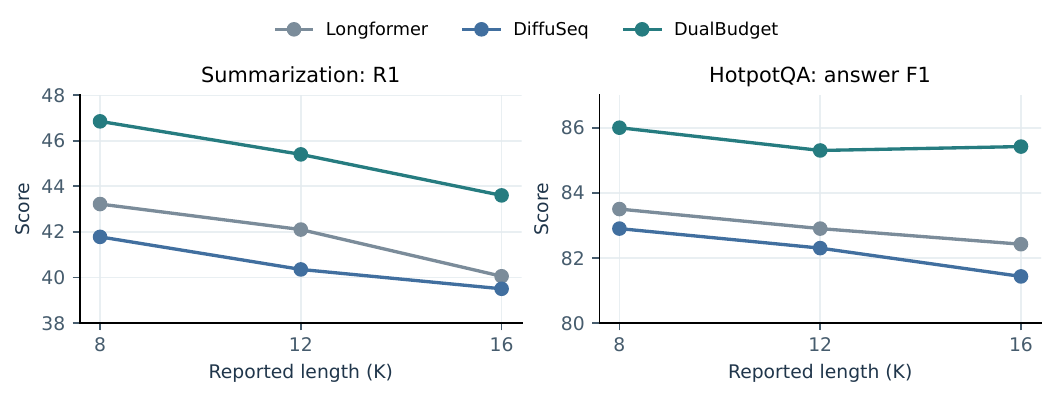}
\caption{Two measured trajectories from the supplied length sweeps: summarization R1 and HotpotQA answer F1. Points reproduce table values; connecting segments guide the eye and do not introduce measurements between tested lengths.}
\label{fig:length}
\end{figure*}

HotpotQA exhibits a more mixed pattern. Answer EM changes from 74.10 at 8K to 72.88 at 16K, while answer F1 changes from 86.00 to 85.42. Answer F1 at 12K is 85.30, so the final increase of 0.12 demonstrates that not every metric is monotonic across the three lengths. Support EM and F1 decrease from 68.00/91.20 to 66.69/90.40. Each trajectory should be read separately.

% Source: tables/hotpotqa_performance_length
\begin{table}[t]
\centering
\small
\setlength{\tabcolsep}{4pt}
\begin{tabular}{lrr}
\toprule
Model & Answer EM/F1 & Support EM/F1\\
\midrule
\multicolumn{3}{c}{8K length}\\
\midrule
Longformer & 72.30 / 83.50 & 66.50 / 90.30\\
DiffuSeq & 71.85 / 82.90 & 66.10 / 89.80\\
\textbf{DualBudget} & \textbf{74.10 / 86.00} & \textbf{68.00 / 91.20}\\
\multicolumn{3}{c}{12K length}\\
\midrule
Longformer & 71.90 / 82.90 & 66.00 / 89.80\\
DiffuSeq & 71.50 / 82.30 & 65.50 / 89.20\\
\textbf{DualBudget} & \textbf{73.50 / 85.30} & \textbf{67.30 / 90.80}\\
\multicolumn{3}{c}{16K length}\\
\midrule
Longformer & 71.21 / 82.42 & 65.11 / 89.50\\
DiffuSeq & 70.91 / 81.43 & 64.60 / 88.51\\
\textbf{DualBudget} & \textbf{72.88 / 85.42} & \textbf{66.69 / 90.40}\\
\bottomrule
\end{tabular}
\caption{HotpotQA scores at 8K, 12K, and 16K. Each pair is exact match / F1.}
\label{tab:hotpotqa_performance_length}
\end{table}

Long-context in-context learning has its own task-dependent failure modes~\cite{li2024long}. The present sweep does not establish a universal 8K boundary for other language models. Nor does the label ``16K'' resolve whether length refers to the input, target, or combined representation. That boundary, truncation policy, output length, and batch construction must accompany a reproducible scaling comparison.

\subsection{Ablating Attention, Steps, and Window Size}
The six configurations in Table~\ref{tab:ablation} preserve the original ablation values. Removing sparse attention changes the three scores by $-1.89/-0.74/-1.48$ relative to the full configuration. Because the dense and sparse kernels, capacities, and training controls are not supplied, this comparison does not independently prove that fewer attention edges cause better quality or faster execution.

% Source: tables/ablation
\begin{table*}[t]
\centering\small
\begin{tabular}{llrrrrr}
\toprule
Setting & Attention & Steps & Window & A & B & C\\
\midrule
Full configuration & Sparse & 2048 & 512 & 44.41 & 18.73 & 39.89\\
Dense attention & Dense & 2048 & 512 & 42.52 & 17.99 & 38.41\\
Fewer steps & Sparse & 1024 & 512 & 43.11 & 18.03 & 39.26\\
More steps & Sparse & 4096 & 512 & 44.71 & 18.55 & 40.20\\
Smaller window & Sparse & 2048 & 256 & 43.80 & 18.22 & 39.65\\
Larger window & Sparse & 2048 & 1024 & 44.40 & 18.66 & 39.92\\
\bottomrule
\end{tabular}
\caption{Six reported ablation configurations. A/B/C preserve the three numeric columns because the source prose's metric labels conflict with the matching summarization baseline. No uncertainty estimates are available.}
\label{tab:ablation}
\end{table*}

Reducing the horizon from 2048 to 1024 changes the scores by $-1.30/-0.70/-0.63$. Increasing it to 4096 changes them by $+0.30/-0.18/+0.31$. The second score becomes lower, so more steps are not uniformly better across the reported triplet. Window 256 changes the scores by $-0.61/-0.51/-0.24$, whereas window 1024 changes them by $-0.01/-0.07/+0.03$. The latter is close to the full setting but does not improve every score.

The ablation prose calls the triplet BLEU/ROUGE/BERTScore, although its baseline exactly matches the aggregate summarization ROUGE triplet. We label the columns A/B/C to retain this unresolved metric identity. This ambiguity does not alter the numerical differences above, but it limits a metric-specific interpretation. Seed-level uncertainty and a clarified evaluator are needed before small changes are treated as reliable improvements.

\subsection{Reported Cost and Novelty}
The inference-cost record lists 0.80 for \method, 1.00 for DiffuSeq, and 1.40 for Longformer. Under one common timing or normalization convention, 0.80 is 20\% below 1.00 and approximately 42.9\% below 1.40. The record therefore ranks \method{} as the lowest-cost entry. It does not support the original prose ordering that called Longformer fastest.

% Source: tables/comparison
\begin{table}[t]
\centering
\small
\setlength{\tabcolsep}{4pt}
\begin{tabular}{lrr}
\toprule
Model & Inference cost & 2-gram novelty\\
\midrule
\textbf{DualBudget} & \textbf{0.80} & \textbf{0.90}\\
DiffuSeq & 1.00 & 0.75\\
Longformer & 1.40 & 0.60\\
\bottomrule
\end{tabular}
\caption{Reported inference cost and 2-gram novelty. The source table does not define the timing unit or normalization.}
\label{tab:comparison}
\end{table}

The same rows report 2-gram novelty of 0.90, 0.75, and 0.60. Their differences are 0.15 and 0.30 in the recorded scale. Higher novelty need not mean greater factuality, useful diversity, or better content coverage; the definition of the reference set and denominator is also needed. A low-cost, high-novelty point should therefore be assessed together with task scores and generation fidelity.

The source table omits the timing unit, batch size, precision, and evaluation schedule. Although an accompanying plot uses a seconds-per-sample label, the underlying timing record is unavailable. We retain the values as reported cost, rather than assert a deployment latency or hardware-independent speedup. No training-time measurements are provided to substantiate the separate 15\% training reduction claim.

% Source: sections/06_discussion
\section{Discussion and Limitations}
\subsection{What the Dual-Budget View Explains}
Context edges control information access, while selected experts control conditional transformation. A disconnected graph cannot be repaired merely by increasing expert count, and global connectivity does not ensure informative expert inputs. The reported scores are compatible with a useful combination, but architecture size, training exposure, conditioning, and decoding remain alternative explanations. The ablations do not separately vary expert activation or the global-token budget. Matched parameter budgets, identical data, measured dispatch, and controlled interventions would isolate these effects.

\subsection{Quality beyond Automatic Overlap}
A scientific summary can match reference phrasing while omitting a limitation or introducing an unsupported causal relation. Answer F1 can improve without better reasoning traces, and novel bi-grams can express useful reformulations or errors. Evaluation should therefore retain support, contradiction, coverage, and discourse judgments alongside overlap scores.

The supplied qualitative asset contains one summary example about tensor models and emergent geometry. It illustrates differences in specificity, but cannot establish an error frequency. A systematic study would retain source documents, generated texts, evaluator versions, and blinded annotations for sampled outputs.

\subsection{Connection to Agent Coordination}
KABB, GAM-Agent, and Markets of Minds~\cite{req_kabb,req_gam,req_markets} motivate an extension in which task-level agents select document spans, negotiate shared capacity, or request specialized reasoning. A coordinator could assign a span priority and a budget, while the denoiser exposes context-edge and expert-activation costs. This connects deliberation about where computation is useful to execution inside each denoising step. No agent coordinator is evaluated in the reported tables.

\subsection{Reproducible Evaluation}
Replication requires four linked descriptions: the model's widths, experts, activation rule, capacity factor, and masks; the diffusion corruption, prediction target, sampler, and time grid; the data splits, source/target lengths, and truncation; and the measurement software, hardware, precision, batch size, evaluator settings, and seeds. Absorbing-state kernels and gate normalization must be included if used.

At each length, an instrumented evaluation should report attention edges, executed expert assignments, dispatch overhead, peak memory, denoiser calls, latency, and throughput. Quality should be compared at matched sampling cost as well as matched update count. Fewer active experts can leave parameter residency unchanged; communication can dominate after attention becomes sparse; additional global positions can erode the linear-cost regime. These measurements would connect the theoretical budgets to practical behavior. They are not additional results contained in the current record.

% Source: sections/07_conclusion
\section{Conclusion}
Sparse text diffusion can be analyzed through separate budgets for context edges and expert activations. The formulation clarifies attention normalization, connectivity, conditional compute, and sampler compatibility. The supplied configuration leads the listed task comparisons through 16K, while retaining length-dependent degradation and mixed ablation effects. Its reported inference-cost ordering is favorable under an unresolved measurement convention. These observations motivate a matched, instrumented evaluation of both budgets, with explicit corruption, routing, and data definitions\label{maintextend}.%

\end{document}